\documentclass[runningheads]{llncs}
\usepackage[T1]{fontenc}
\usepackage{subcaption}
\usepackage{multirow}
\usepackage{algorithm2e}   
\usepackage{cite}
\usepackage{amsmath,amssymb,amsfonts}
\usepackage{graphicx}
\usepackage{textcomp}
\usepackage{xcolor}
\usepackage[font=small,labelfont=bf]{caption}
\usepackage{amsmath}

\begin{document}
\title{iPac: Incorporating Intra-image Patch Context into Graph Neural Networks for Medical Image Classification}
%
%
\author{Usama Zidan\inst{1} \and
Mohammed Medhat Gaber\inst{2} \and
Mohammed M. Abdelsamea\inst{3}\thanks{Corresponding author: \email{m.abdelsamea@exeter.ac.uk}}}

\authorrunning{Zidan et al.}

\institute{Nuffield Department of Surgical Sciences, University of Oxford, UK
\and
School of Computing and Digital Technology, Birmingham City University, UK
\and
Department of Computer Science, University of Exeter, UK}

\maketitle              
\begin{abstract}
Graph neural networks have emerged as a promising paradigm for image processing, yet their performance in image classification tasks is hindered by a limited consideration of the underlying structure and relationships among visual entities. This work presents iPac, a novel approach to introduce a new graph representation of images to enhance graph neural network image classification by recognizing the importance of underlying structure and relationships in medical image classification. iPac integrates various stages, including patch partitioning, feature extraction, clustering, graph construction, and graph-based learning, into a unified network to advance graph neural network image classification. By capturing relevant features and organising them into clusters, we construct a meaningful graph representation that effectively encapsulates the semantics of the image. Experimental evaluation on diverse medical image datasets demonstrates the efficacy of iPac, exhibiting an average accuracy improvement of up to  5\% over baseline methods. Our approach offers a versatile and generic solution for image classification, particularly in the realm of medical images, by leveraging the graph representation and accounting for the inherent structure and relationships among visual entities.

\keywords{Graph neural networks (GNNs) \and medical image classification \and clustering.}
\end{abstract}
\section{Introduction}

Image classification is a core task in computer vision, involving the assignment of labels to images based on their content. Traditional deep learning methods such as convolutional neural networks (CNNs) and transformers have shown significant success by leveraging the grid or sequence structure of images. However, there is growing interest in using graph neural networks (GNNs) for image classification due to their ability to capture relational information and handle nongrid structured data. GNNs are particularly beneficial for medical images, where objects of interest often have irregular shapes and overlapping characteristics.

Several state-of-the-art methods have explored the use of GNNs for image classification, particularly in the medical domain. Recent approaches have tackled the challenge of representing images as graphs while preserving spatial and semantic information. Methods such as superpixel over-segmentation \cite{gan2022image} and region proposals have been employed to convert images into graph representations. The efficacy of Graph Attention Networks (GATs) for image classification has been demonstrated in \cite{gnn_super}, achieving high performance on standard datasets, but noting potential issues with information loss and memory usage during graph generation. Hierarchical GNN \cite{du2019zoom} was introduced to detect abnormal lesions in mammograms, combining CNNs for feature extraction with GATs for node and graph classification, achieving state-of-the-art performance on the INbreast dataset. Similarly, a synergic GNN for medical image classification was proposed in \cite{yang2019classification}, enhancing accuracy by combining features from different Graph Convolutional Network (GCN) variants. In \cite{bai2022applying}, a GNN-based model was proposed in a way to make use of a self-attention mechanism to classify digital breast tomosynthesis images and has shown superior performance compared to traditional CNN methods. Additionally, GNNs have been leveraged for detecting and segmenting lymph node tumors in oncology imaging, demonstrating significant improvements over CNN-based models \cite{chao2020lymph}. An attention-guided deep GNN has been applied to analyze the progression of Alzheimer's disease, achieving superior classification accuracy by processing brain network data from MRI images \cite{ma2020attention}. Furthermore, integration of a GNN module with a 3D UNet to segment tree-like structures on chest CT has showcased the benefits of graph neighborhood connectivity \cite{garcia2019joint}. These recent methods highlight the advancements in using GNNs for complex image processing tasks, addressing challenges in capturing semantic and structural information of medical images.

In medical imaging, accurate classification often requires capturing intricate details of cell/tissue morphology and spatial relationships, which traditional pixel-based methods may miss. To address this, we propose iPac, a novel approach that converts images into graphs that can model the relationships among different clusters of patches, to enhance classification accuracy using GNNs.  iPac is evaluated on various medical image classification tasks, including histological images, skin lesion diagnosis, and retinal imaging. iPac outperforms state-of-the-art methods using GNNs on regular grids or similarly sized superpixels, demonstrating the effectiveness of our image-to-graph conversion and GNN-based classification approach.

\begin{figure*}[t]
    \centering
    \includegraphics[width=\textwidth]{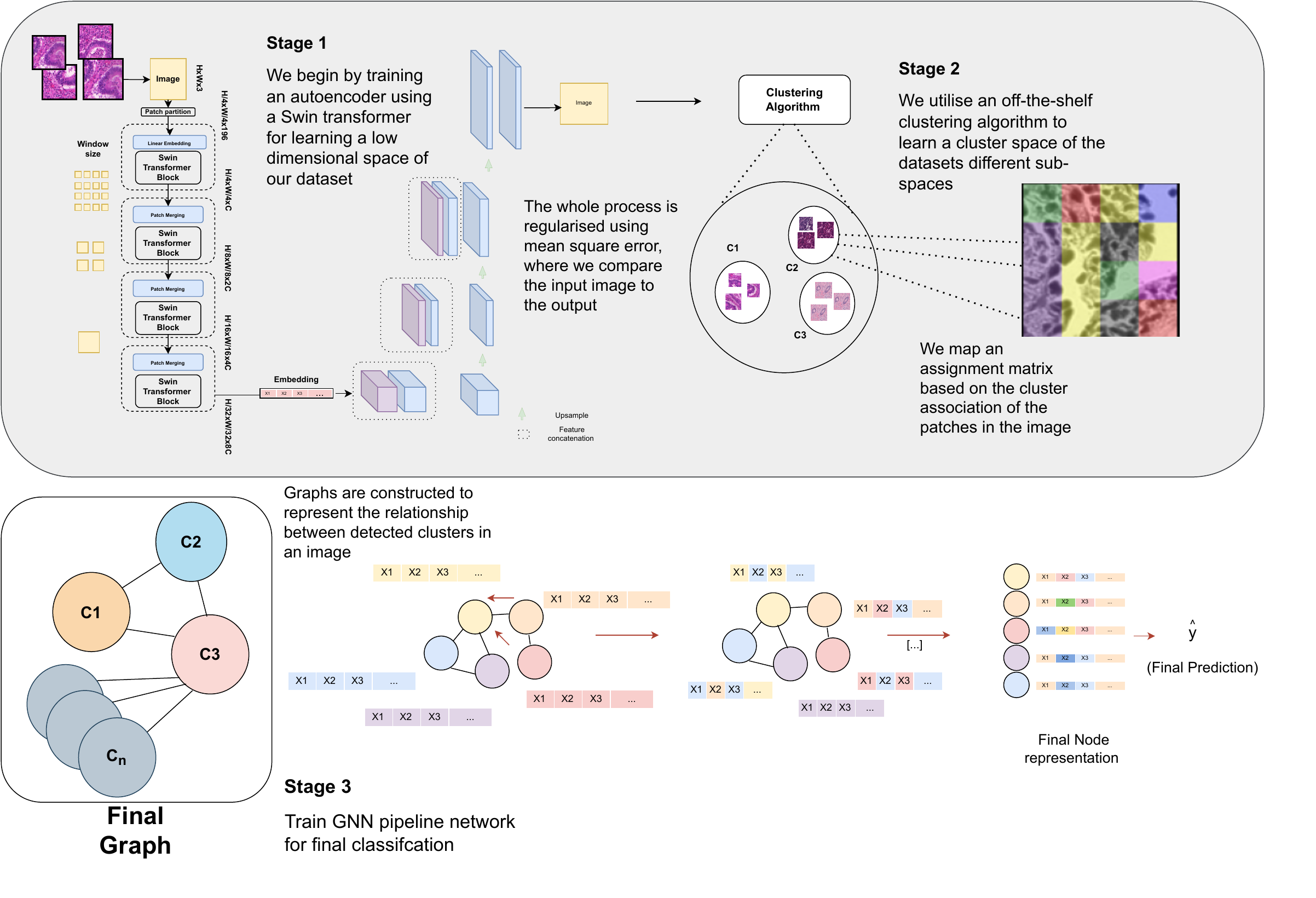}
    \caption{iPac consists of three main stages: image patching, patch encoding and clustering, and graph formation. We divide an input image into several patches, encode them using a Swin transformer autoencoder, and cluster them. A GNN is constructed to represent the detected clusters in an image with layers iterating over the node representation of each cluster, taking into account the edge weights that represent the inherent relationship between patch and cluster pairs. }
    \label{fig:main_arch}
\end{figure*}

\section{Methodology}

Our proposed approach, Figure \ref{fig:main_arch}, is a multistage method to convert an image into a graph representation, addressing limitations of pixel-based approaches and super-pixel representations of current mode. Initially, the input image is partitioned into a fixed grid of patches. We train a Swin Transformer-based autoencoder to encode these patches into high-dimensional embeddings that capture spatial and semantic information. These embeddings are then clustered to form a compact set, each represented by its centroid. Let $x_{i,j}$ be the $j$th patch in the $i$th image, with $j \in \{1,\dots,P\}$, and $P$ being the total number of patches. We train an autoencoder with encoder $f_{\theta}$ and decoder $g_{\phi}$ to reconstruct $x_{i,j}$ from a compressed vector representation $z_{i,j}$, optimizing the mean squared error. The complete pipeline is outlined in Algorithm~\ref{algo:patchgraphnet}.

\begin{algorithm}[htbp]
\SetAlgoLined
\KwData{Set of images $\mathcal{I}$}
\KwResult{Graph representations for each image in $\mathcal{I}$}
\ForEach{image $I$ in $\mathcal{I}$}{
    $P \gets$ getPatches($I$) \tcp*{Split image into patches}
    $E \gets$ Encoder($P$) \tcp*{Obtain embeddings for patches}
    $C \gets$ detectClusters($E$) \tcp*{Detect clusters using clustering algorithm}
    $G \gets$ constructGraph($C$, $P$) \tcp*{Construct Graph}
    \textbf{use} $G$ \textbf{for graph-based learning and image classification}\;
}
\caption{iPac algorithm}
\label{algo:patchgraphnet}
\end{algorithm}

\begin{figure}[tbh]
    \centering
    \includegraphics[width=.9\textwidth]{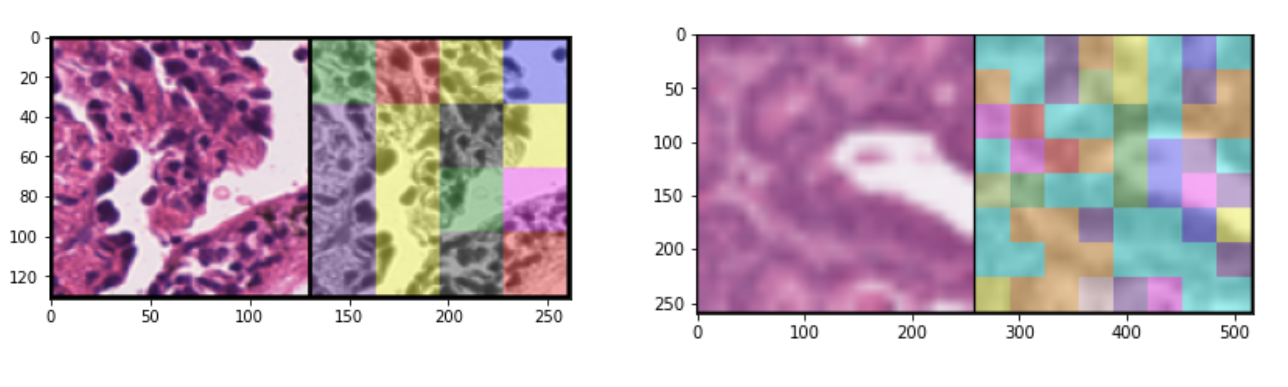}
    \caption{An example of cluster assignment on the input image. Each patch is processed through an autoencoder and assigned a cluster using a clustering algorithm. }
    \label{fig:cluster_assign}
\end{figure}

After training an encoder, we encode patches from the input image. The method for determining the number of patches per image is detailed in Section \ref{ablation_tests}. Each patch, now represented in a high-dimensional space, is clustered using an off-the-shelf clustering algorithm, such as \textit{k-means}, to form a high-level representation of the objects in the image.
We construct an adjacency matrix $A$ based on the spatial distribution of these clustered patches (Figure \ref{fig:cluster_assign}). Formally, $A$ is a $C \times C$ matrix, where $C$ is the number of clusters. Elements of $A$ are defined as:
\begin{equation}
A_{ij} = \begin{cases}
\frac{n_{ij}}{n_{i,\cdot}}, & \text{if } n_{ij} > 0 \\
0, & \text{otherwise}
\end{cases}
\end{equation}
where $n_{ij}$ is the number of patches from cluster $j$ adjacent to patches of cluster $i$, and $n_{i,\cdot}$ is the total number of patches adjacent to patches of cluster $i$. The image is divided into patches where each patch is assigned to a cluster based on clustering performed on the entire dataset. We aggregate each cluster's patches as a node representation. 


In the second stage, we construct a graph in which the nodes represent clusters of patches identified within the image. To connect these nodes, we analyse the spatial relationships between patches belonging to different clusters. Specifically, we examine how frequently patches from one cluster are adjacent to patches from another cluster within the image. The more often these patches are neighbours, the stronger the connection between their respective cluster nodes in the graph. This adjacency-based weighting of edges ensures that the graph reflects meaningful spatial relationships present in the image. Algorithm \ref{algo:constructgraph} showcases the exact steps in graph construction.

\begin{algorithm}[t]
\SetAlgoLined
\KwData{Cluster labels $C$, Encoded image patches $P$}
\KwResult{Graph representation $G$}
$G \gets$ initializeEmptyGraph() \tcp*{Initialize an empty graph}

\ForEach{cluster $c$ in $C$}{
    addNode($G$, $c$) \tcp*{Add a node to the graph representing the cluster}
}

\ForEach{patch $p$ in $P$}{
    $c_p \gets$ getClusterLabel($p$) \tcp*{Retrieve the cluster label of the patch}
    $N \gets$ getNeighboringPatches($p$) \tcp*{Get neighboring patches of $p$}
    
    \ForEach{neighbor patch $q$ in $N$}{
        $c_q \gets$ getClusterLabel($q$) \tcp*{Retrieve the cluster label of the neighbor patch}
        
        \If{$c_p \neq c_q$}{
            $w_{pq} \gets$ calculateEdgeWeight($c_p, c_q$) \tcp*{Calculate edge weight based on cluster occurrences}
            addEdge($G$, $c_p, c_q$, $w_{pq}$) \tcp*{Add edge to graph with weight}
        }
        \ElseIf{$c_p = c_q$}{
            addSelfEdge($G$, $c_p$) \tcp*{Add self-edge}
        }
    }
}
\Return $G$
\caption{Construct Graph}
\label{algo:constructgraph}
\end{algorithm}

Finally, in the third stage, we leverage edge-based GNN layers for image classification, using the graph structure to improve accuracy. The use of GNNs allows us to leverage the structure of the graph representation to improve classification performance and has been shown to be effective for a variety of image processing tasks \cite{gnn_super}. Figure \ref{fig:cluster_assign} shows an example of cluster assignment in image patches. Each patch is associated with a cluster, facilitating the formation of a graph. 

In the iPac model, an extension to traditional GCNs called edge graph convolution is employed to capture additional information from the graph structure. Edge graph convolution operates by considering not only the node representations but also the edge attributes in the graph.
The edge graph convolution is formulated as follows:

\begin{equation}
\begin{aligned}
h_i^{(l+1)} &= \sigma \left( \sum_{j \in \mathcal{N}(i)} \frac{1}{\sqrt{|\mathcal{N}(i)| \cdot |\mathcal{N}(j)|}} \cdot \Theta^{(l)} \cdot h_j^{(l)} \right. \\
&\quad \left. + \sum_{j \in \mathcal{E}(i)} \frac{1}{\sqrt{|\mathcal{E}(i)| \cdot |\mathcal{E}(j)|}} \cdot \Phi^{(l)} \cdot e_{ij}^{(l)} \right)
\end{aligned}
\end{equation}

where $h_i^{(l+1)}$ represents the updated feature representation of node $i$ at layer $l+1$, $\mathcal{N}(i)$ denotes the set of neighboring nodes of node $i$, $\mathcal{E}(i)$ represents the set of edges connected to node $i$, $\Theta^{(l)}$ and $\Phi^{(l)}$ are learnable weight matrices at layer $l$, and $e_{ij}^{(l)}$ denotes the edge feature between nodes $i$ and $j$ at layer $l$.

By incorporating the edge features $e_{ij}^{(l)}$ into the aggregation process, the model captures not only the node-level information but also the relationships between nodes through their associated edges. Considering edge features enhances the ability to capture fine-grained structural information and achieve more expressive graph representations, leading to improved graph classification performance.

After applying the edge graph convolution, the next step in iPac is message passing. Message passing enables information exchange between nodes and helps propagate relevant information across the graph. This process allows nodes to gather and update their feature representations based on the information received from their neighboring nodes. The message passing operation in iPac is defined as follows:

\begin{equation}
m_{ij}^{(l+1)} = \phi^{(l)} \left( h_i^{(l+1)}, h_j^{(l+1)}, e_{ij}^{(l)} \right)
\end{equation}

where $m_{ij}^{(l+1)}$ represents the message passed from node $i$ to node $j$ in layer $l+1$, and $\phi^{(l)}$ is a learnable function that incorporates the node features $h_i^{(l+1)}$ and $h_j^{(l+1)}$, as well as the edge feature $e_{ij}^{(l)}$.

Finally, the feature representations obtained after the message passing phase are fed into a fully connected layer with softmax activation. This layer produces the probability distribution over the target classes, allowing us to classify the graph.

\begin{equation}
\hat{y}_i = \text{softmax} \left( \sum_{j \in \mathcal{N}(i)} m_{ij}^{(L)} \right)
\end{equation}

where $\hat{y}_i$ represents the predicted class probabilities for graph $i$, $\mathcal{N}(i)$ denotes the set of neighboring nodes of node $i$, and $L$ represents the total number of layers in the network.


In iPac, training involves first using an autoencoder to learn cluster representations from the entire dataset. During inference, these pre-trained clusters are detected in individual test images, and a graph is constructed with nodes representing these clusters.  This approach leverages this graph representation to encode both local and global information, thereby enhancing performance and interpretability in image classification tasks. Locally, the graph captures the detailed relationships between patches within an image, ensuring that fine-grained features are preserved and used. Globally, it models the interactions between these patches through cluster adjacency, providing a high-level abstraction of the image structure. This is particularly beneficial for medical images, where complex structures require nuanced understanding. For example, in histological images, the clustering of pixels into meaningful groups reduces noise and highlights distinctive features, while the graph structure captures the broader context and interactions between different biological entities, thereby improving the accuracy and reliability of classification.

\section{Results and Discussion}



\begin{table*}[!ht]
\centering
\caption{Comparison of Results on Evaluation Datasets. We compare our proposed approach to a GNN-based approach along with recent SOTA methods, showing Area Under the Curve (AUC) and Accuracy (ACC) results.}
\label{tab:results}
\resizebox{\textwidth}{!}{%
\begin{tabular}{|l|cc|cc|cc|cc|}
\hline
\textbf{Methods} &
  \multicolumn{2}{c|}{\textbf{PathMNIST}} &
  \multicolumn{2}{c|}{\textbf{DermaMNIST}} &
  \multicolumn{2}{c|}{\textbf{BreastMNIST}} &
  \multicolumn{2}{c|}{\textbf{RetinaMNIST}} \\ \cline{2-9}
 &
 \textbf{AUC} & \textbf{ACC} &
 \textbf{AUC} & \textbf{ACC} &
 \textbf{AUC} & \textbf{ACC} &
 \textbf{AUC} & \textbf{ACC} \\ \hline

\textbf{Baseline GNN} & 98.0\% & 90\% & 90.2\% & 71\% & 83.6\% & 80\% & 71.6\% & 51\% \\ \hline
\textbf{ResNet-18 (224) \cite{resnet}} & 98.0\% & 90.9\% & 90.2\% & 71.4\% & 85\% & 82\% & 72.6\% & 52.8\% \\ \hline
\textbf{ResNet-50 (224) \cite{resnet}} & 98.0\% & 89.2\% & 91\% & 71.9\% & 84.6\% & 83.4\% & 71.6\% & 52.4\% \\ \hline
\textbf{MedVit-T \cite{manzari2023medvit}} & 99.4\% & 93.8\% & 91.4\% & 76.8\% & 93.4\% & 89.6\% & 75.2\% & 53.4\% \\ \hline
\textbf{MedMamba-T \cite{yue2024medmamba}} & 99.7\% & 95.3\% & 91.7\% & 77.9\% & 82.5\% & 85.3\% & 74.7\% & 54.3\% \\ \hline
\textbf{iPac} & 99.4\% & 91.4\% & 92.5\% & 73.4\% & 91.9\% & 85.4\% & 76.6\% & 54.6\% \\ \hline

\end{tabular}%
}
\end{table*}

\subsection{Datasets}

During our training and evaluation process, we used a list of four medical datasets of various modalities. These datasets represent various medical imaging workflows in which diagnosis is heavily based on imaging interpretation. The four datasets from MedMNIST are 2D to evaluate the performance of our approach. These datasets include the retinal, skin, pathology, and breast datasets. MedMNIST is a lightweight large-scale benchmark for the classification of 2D and 3D biomedical images \cite{yang2023medmnist}. It consists of 12 preprocessed 2D datasets and 6 preprocessed 3D datasets from selected sources covering primary data modalities (e.g., X-Ray, Ultrasound, CT, Electron Microscope), diverse classification tasks (binary/multiclass) and data scales (from 1000 to 100,000). Using these specific datasets from MedMNIST, we were able to test iPac's performance in medical imaging.

\subsection{Performance Evaluation}

The results of our experiments are shown in Table~\ref{tab:results}. iPac is compared to variants state-of-the-art methods, including a baseline GNN superpixel-based methods representing the common approach to classifying images using GNNs. iPac outperformed the baseline GNN methods in all datasets, achieving higher accuracy. Specifically, our method achieved an accuracy of 85. 4\%, 54. 63\%, 91. 4\% and 73\% in BreastMNIST, RetinaMNIST, PathMNIST, and DermaMNIST, respectively. The baseline GNN method achieved an accuracy of 80\%, 51\%, 90\%, and 71\% in the same datasets. 

Table \ref{tab:results} reports the performance comparison of iPac with previous state-of-the-art methods in terms of AUC and ACC in each MedMNIST dataset. Compared to recent MedViT-T, the AUC of iPac in DermaMNIST is 1. 1\% higher, indicating that iPac maintains a clear advantage in image-based classification tasks for DermaMNIST. Furthermore, compared to the MedMamba model in BreastMNIST, iPac shows competitive performance, achieving a similar ACC and a 9. 4\% higher AUC. Overall, the iPac model effectively improves the performance of medical image classification tasks in the MedMNIST benchmark, especially for PathMNIST, DermaMNIST, BreastMNIST, and RetinaMNIST. The performance of the iPac model on MedMNIST establishes a new baseline for GNN based methods and achieves state-of-the-art results, making it a highly effective tool for medical image classification tasks.



By clustering image patches into graph nodes, iPac preserves both local details and the global context essential for accurate medical diagnosis. Unlike superpixel graphs, iPac retains key visual patterns and relationships. Operating on patches rather than individual pixels allows iPac to capture higher-level anatomical features, enhancing pathology detection accuracy. iPac's ability to model long-range spatial relationships and richer anatomical representations offers performance advantages over CNNs, which primarily focus on local features. Additionally, iPac's conversion of images into sparse graphs enables efficient training on full-sized medical images without losing finer details critical for accurate classification.


\subsection{Ablation Tests}
\label{ablation_tests}
We performed studies on different parameters that governed the graph formation and removing and adding different layers of the model. The patch size ranged from 8 to 112, ensuring at least 4 patches per image given a fixed image size of 224$\times$224 \cite{yang2023medmnist}. The number of clusters varied from 4 to 224, covering a wide range of configurations. For each trial, we trained an autoencoder on images of the chosen patch size and applied the \textit{k-means} algorithm to determine cluster centroids. The resulting graphs were then used for classification and the accuracy of the validation was recorded for each configuration.
\vspace{-30pt} 

\begin{table}
\centering
\caption{Average accuracy (with standard deviation) of different layer types.}
\label{tab:layer_type_accuracy}
\begin{tabular}{|l|c|}
\hline
\textbf{Layer Type} & \textbf{Accuracy (Mean $\pm$ Std)} \\
\hline
gatconv & 0.8782 $\pm$ 0.0121 \\
gcnconv & 0.8882 $\pm$ 0.0076 \\
ginconv & 0.8694 $\pm$ 0.0237 \\
sageconv & 0.9221 $\pm$ 0.0072 \\
\hline
\end{tabular}
\end{table}

\begin{table}
\centering
\caption{Average accuracy across various inner Dimensions.}
\label{tab:inner_dim_accuracy}
\begin{tabular}{|l|c|c|c|}
\hline
\textbf{Layer Type} & \textbf{128} & \textbf{256} & \textbf{512} \\
\hline
gatconv & 0.8910 $\pm$ 0.0075 & 0.8843 $\pm$ 0.0082 & 0.8713 $\pm$ 0.0107 \\
gcnconv & 0.8939 $\pm$ 0.0036 & 0.8918 $\pm$ 0.0041 & 0.8825 $\pm$ 0.0066 \\
ginconv & 0.8747 $\pm$ 0.0077 & 0.8657 $\pm$ 0.0216 & 0.8529 $\pm$ 0.0304 \\
sageconv & 0.9270 $\pm$ 0.0043 & 0.9243 $\pm$ 0.0060 & 0.9230 $\pm$ 0.0071 \\
\hline
\end{tabular}
\end{table}


\begin{table}
\centering
\caption{Average accuracy with different configurations of number of MLP layers.}
\label{tab:num_layers_accuracy}
\begin{tabular}{|l|c|c|c|}
\hline
\textbf{Layer Type} & \textbf{4 Layers} & \textbf{6 Layers} & \textbf{12 Layers} \\
\hline
gatconv & 0.8909 $\pm$ 0.0078 & 0.8763 $\pm$ 0.0110 & 0.8794 $\pm$ 0.0071 \\
gcnconv & 0.8947 $\pm$ 0.0030 & 0.8889 $\pm$ 0.0056 & 0.8846 $\pm$ 0.0072 \\
ginconv & 0.8786 $\pm$ 0.0025 & 0.8735 $\pm$ 0.0085 & 0.8415 $\pm$ 0.0326 \\
sageconv & 0.9303 $\pm$ 0.0026 & 0.9275 $\pm$ 0.0016 & 0.9165 $\pm$ 0.0043 \\
\hline
\end{tabular}
\end{table}

\vspace{30pt} 



\subsubsection{GCN Architecture Comparison} In addition to hyperparameter optimization, we performed a comprehensive evaluation of various GCN architectures to assess their impact on model performance. Our evaluation encompassed different types of GCN layer, number of layers, dropout rates, inner dimension sizes, and multilayer perceptrons (MLPs) configurations at the start and end of the model.

We examined multiple types of GNN layer including the standard GCN, Chebyshev \cite{he2022convolutional}, and GraphSAGE layers \cite{hamilton2017inductive}. The number of layers in our evaluations varied from one to five, exploring the optimal layer depth for our datasets. Dropout rates ranging from 0.1 to 0.8 were tested to mitigate overfit during training.

The inner dimension size of GCN layers, crucial for feature extraction and representation, was evaluated across dimensions from 128 to 512 to determine the most effective size for our datasets. In addition, we explore the impact of MLPs at both the input and output ends of the model, varying the number of MLPs from 4, 6, and 12 layers.

This systematic exploration aimed to identify the optimal GCN architecture configuration that maximizes classification accuracy across our datasets, providing insight into effective model design for graph-based medical image analysis.

Based on our results, the type of GraphSAGE layer (sageconv) consistently outperformed other types in most evaluations, with an average accuracy of \(0.9221 \pm 0.0072\) (Table \ref{tab:layer_type_accuracy}). The GCN (gcnconv) layer type also showed competitive performance, achieving an average accuracy of \(0.8882 \pm 0.0076\), though it was not as consistently as sageconv. In contrast, the Graph Isomorphism Network (ginconv) and Graph Attention Network (gatconv) layer types generally showed lower accuracy, with ginconv showing an average accuracy of \(0.8694 \pm 0.0237\) and gatconv \(0.8782 \pm 0.0121\).

Our analysis revealed notable trends regarding inner dimensions and the number of layers in GNN architectures. For inner dimensions, we observed that increasing dimension size generally led to a decrease in performance across most layer types (Table \ref{tab:inner_dim_accuracy}). Furthermore, the evaluation of different numbers of layers indicated that accuracy tended to decrease with an increase in the number of layers (Table \ref{tab:num_layers_accuracy}). These observations suggest that overly complex feature spaces and more MLP layers may introduce unnecessary complexity that lead to overfitting and poor generalization. These findings emphasize the critical role of selecting the optimal parameters to maintain a balance between model depth and performance.


\subsubsection{Hyperparameter Optimization}

\begin{figure}[ht]
    \centering
        \includegraphics[width=.9\textwidth]{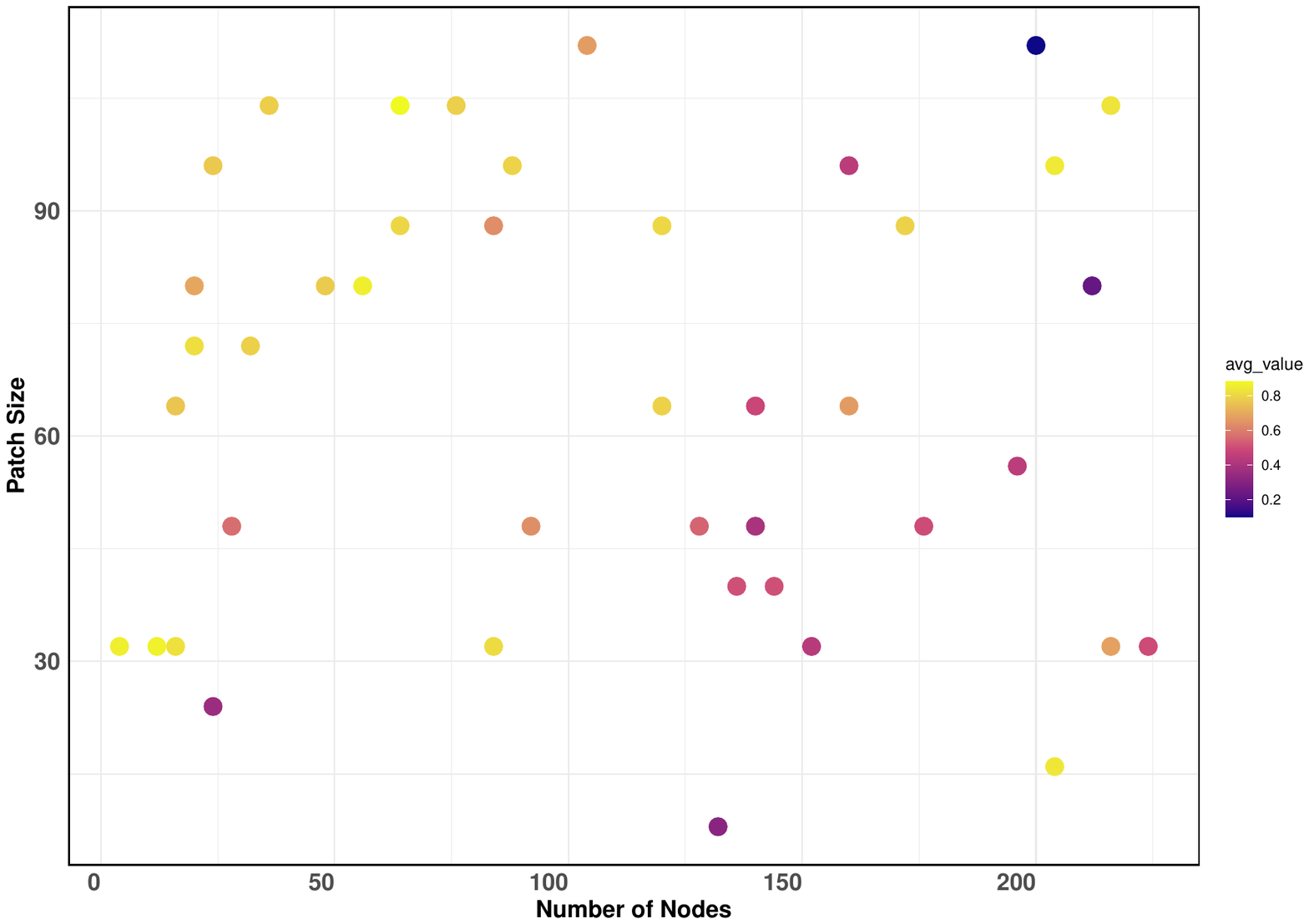 }
        \caption{Average accuracy for each patch size and number of nodes/clusters pairs.}
        \label{fig:scatter-all}
    \end{figure}

To determine the optimal hyperparameters, we performed hyperparameter optimization using the Optuna library \cite{akiba2019optuna}, employing a genetic algorithm approach. Key parameters such as the number of graph clusters and patch size were tuned on a validation set, optimizing for cross-entropy loss. Figure \ref{fig:scatter-all} shows the results of hyperparameter optimisation for different image classification datasets using the Optuna library. We observe a trend of higher performance achieved with a lower number of clusters in the graph.  Notably, models with fewer clusters consistently achieved higher validation accuracy.  This suggests that reducing the complexity of the graph by employing fewer clusters can enhance the accuracy of iPac on some datasets. 

\begin{figure}[ht]
    \centering
    \begin{subfigure}[b]{0.4\textwidth}
        \includegraphics[width=\textwidth]{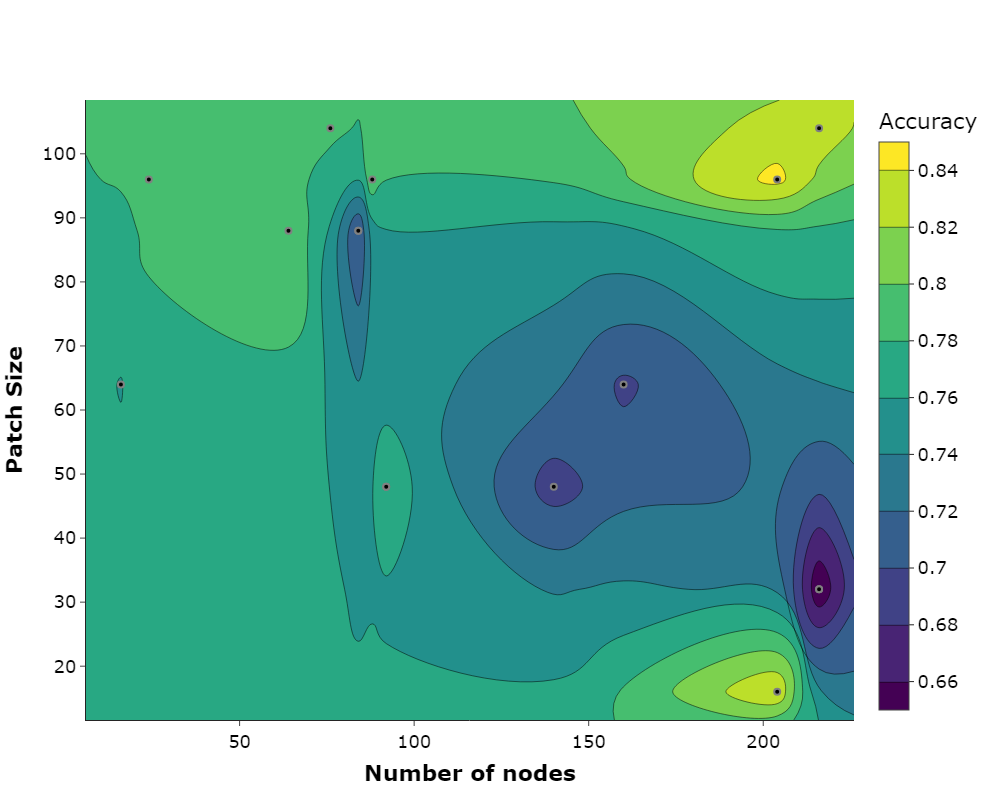}
        \caption{BreastMNIST1}
        \label{fig:dataset1_contour}
    \end{subfigure}
    \begin{subfigure}[b]{0.4\textwidth}
        \includegraphics[width=\textwidth]{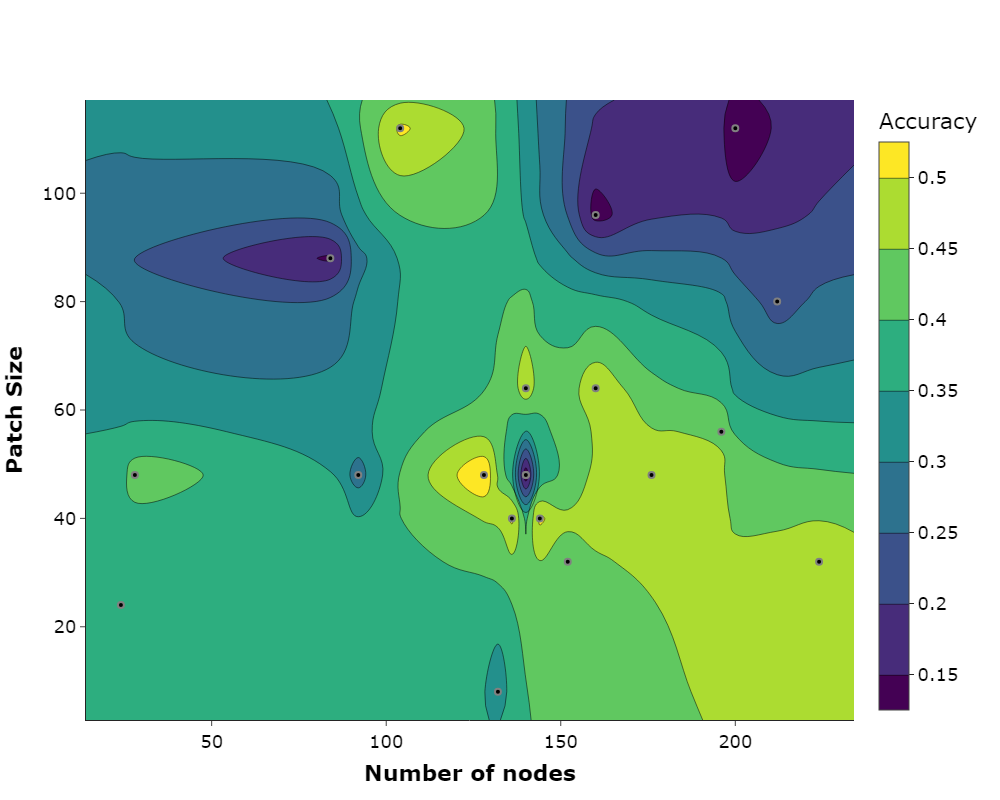}
        \caption{RetinalMNIST}
        \label{fig:dataset3_contour}
    \end{subfigure}
    \caption{Contour plots of search space with accuracy overlaid. Each dataset shows a different landscape portraying how different medical images require different tuned parameters.}
    \label{fig:contour_plots}
\end{figure}
For the BreastMNIST dataset, figure \ref{fig:dataset1_contour}, the highest accuracy of 84\% was achieved with 204 nodes and a patch size of 96. However, the second highest accuracy of 83\% was achieved with the same number of clusters but a smaller patch size of 16. This suggests that for this dataset, the patch size has a significant impact on performance, while the number of clusters may not be as critical.
The RetinaMNIST dataset, figure \ref{fig:dataset3_contour}, had the lowest overall accuracy among the datasets, with a maximum accuracy of 54\% achieved with 128 nodes and a patch size of 48. This shows the complexity of the retinal image diagnosis.  In contrast to other trials, higher cluster numbers coupled with smaller patch sizes yielded better performance on the more challenging RetinaMNIST dataset.

In the BreastMNIST dataset, models with fewer nodes, such as those with 12, 16, or 24 clusters, consistently achieved higher validation accuracy. Conversely, models with a higher number of clusters, such as those with 160 or 200 clusters, achieved lower validation accuracy.
In trials of the patch size, we see a range of optimal patch sizes across the different datasets. For example, for BreastMNIST, the best validation accuracy was achieved with a patch size of 96 or 104. In contrast, for RetinaMNIST, a lower patch size of 8 or 24 yielded the best results. In general, there was a trend of increasing validation accuracy with increasing patch size for most datasets.

\begin{figure}[!h]
    \centering
    \includegraphics[width=.45\textwidth]{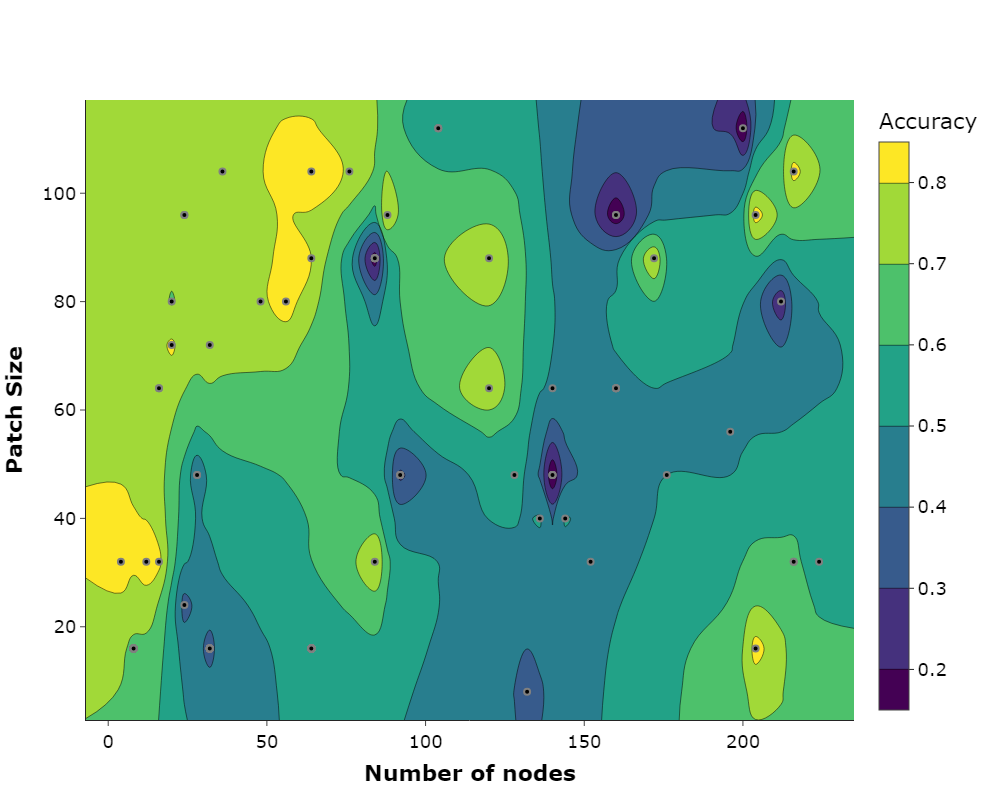}
    \caption{Contour plot of search space complied over all trials in each dataset optimisation.}
    \label{fig:contcawour_all}
\end{figure}

Overall, these results suggest that a lower number of clusters, Figure \ref{fig:contcawour_all}, in the graph may be more beneficial to achieve higher accuracy in graph neural networks for image classification tasks. The optimal patch size may vary depending on the dataset and the task at hand. Our thorough hyperparameter optimisation provides valuable guidelines for configuring iPac across different medical imaging tasks. The key trends observed were: fewer clusters and moderate patch sizes work best for simpler datasets like BreastMNIST, while more clusters and smaller patches suit challenging tasks like RetinaMNIST. Based on our results, starting with graph representations of around 50-100 clusters and patch sizes between 32-96 pixels. These settings provide a good balance of performance and efficiency. Furthermore, extensive hyperparameter search is advised as the optimal settings can be dataset-dependent. By tuning the graph construction, iPac can be adapted to maximise classification accuracy for diverse medical imaging applications.


\section{Conclusion}

In conclusion, iPac, a novel approach for image classification based on graph neural networks, involves three main steps: patch extraction, clustering, and graph construction. Through extensive experiments, iPac demonstrated superior performance compared to existing state-of-the-art methods in image classification. It introduces a graph-based representation of images, where patches are extracted and clustered, enabling capture of both local and global features. Constructing a graph with clusters as nodes and patch similarities as edges effectively leverages graph neural networks for image classification. iPac offers a promising avenue for applying GNNs to image data, suggesting potential extensions such as exploring different clustering algorithms and integrating attention mechanisms.

\bibliographystyle{splncs04}
\bibliography{ICONIP}

\end{document}